# Veagle: Advancements in Multimodal Representation Learning

*Rajat Chawla*   *Arkajit Datta*   *Tushar Verma*   *Adarsh Jha*
*Mukunda NS*   *Ishaan Bhola*   *Ayush Vatsal*   *Sukrit Chaterjee*   *Anmol Gautam*

SuperAGI

## Abstract

Lately, researchers in artificial intelligence have been really in-terested in how language and vision come together, giving rise to the development of multimodal models that aim to seamlessly integrate textual and visual information. Multimodal models, an extension of Large Language Models (LLMs), have exhib- ited remarkable capabilities in addressing a diverse array of tasks, ranging from image captioning and visual question an- swering (VQA) to visual grounding. While these models have showcased significant advancements, challenges persist in accu- rately interpreting images and answering the question, a com- mon occurrence in real-world scenarios. This paper introduces a novel approach to enhance the multimodal capabilities of ex-isting models. In response to the limitations observed in cur- rent Vision Language Models (VLMs) and Multimodal Large Language Models (MLLMs), our proposed model Veagle, in- corporates a unique mechanism inspired by the successes and insights of previous works. Veagle leverages a dynamic mech- anism to project encoded visual information directly into the language model. This dynamic approach allows for a more nu- anced understanding of intricate details present in visual con- texts. To validate the effectiveness of Veagle, we conduct com- prehensive experiments on benchmark datasets, emphasizing tasks such as visual question answering and image understand-ing. Our results indicate a improvement of 5-6 % in perfor- mance, with Veagle outperforming existing models by a no- table margin. The outcomes underscore the model's versatil- ity and applicability beyond traditional benchmarks. Further- more, we make our code and models openly accessible to the research community, fostering collaboration and further explo- ration in the evolving landscape of multimodal AI. The code repository, along with detailed documentation, can be found at https://github.com/superagi/Veagle

**Index Terms**: MultiModal, Large language models, vision encoder, vision abstractor, Q-former, Image-Text multimodality

## 1. Introduction

In recent years, the surge of interest in Large Language Models(LLMs) has reshaped the landscape of natural language under- standing, a significant surge in the examination and application of Multimodal Large Language Models (MLLMs) has been ob- served. Allowing models to harness various modalities such as text, images, videos, and voice, MLLMs have become vi- tal in the creation of adaptable all-purpose assistants. Despite their impressive generalization abilities across a wide spec- trum of tasks and the development of Vision Language Mod- els (VLMs) which incorporate LLMs with visual understanding competence, contemporary models encounter challenges in in- terpreting embedded text within images. This limitation is the focal point of this research as images inclusive of text are preva- lent in our everyday lives and comprehending such content is imperative for human visual perception.

Our research presents a new way of doing things by com-bining learned query embeddings with additional visual assis-tance. This method uses encoded patch embeddings to deal with the limitations of information that language models typi-cally get from images. As a result, it enhances how well a model can understand and perceive the relationship between text and images. Our model, called Veagle, starts by using a pre-trained vision encoder and language model. We train it in two stages to avoid forgetting what it already knows and make training less complicated, ultimately making the model more effective. We tested the model using standard Visual Question-Answering (VQA) benchmarks and protocols for evaluating images with a lot of text. Our Veagle model significantly improves the un- derstanding and perception of the relationship between text and images, outperforming traditional benchmarks in addressing the challenges of comprehending embedded text within images.

In this research, we present Veagle, an innovative model that represents a significant leap forward in the field of multi-modal learning and interpretation. At the heart of Veagle is the incorporation of an enhanced version of the BLIVA [1] archi-tecture, where cutting-edge components synergize to amplify its capabilities. Notably, we integrate a superior vision abstrac-tor sourced from mPlugOwl[2], enhancing the model's visual processing capabilities. This vision abstractor, combined with Q-Former from InstructBLIP[3] and Mistral[4], a Large Lan- guage Model (LLM), creates a powerful synergy, resulting in a substantial improvement in the overall accuracy of the model. A crucial aspect of our methodology is the inclusion of a vision encoder, meticulously trained by mPlugOwl[2]. This encoder plays a pivotal role in extracting high-level visual features from images, thereby enabling Veagle to capture essential visual in- formation for accurate interpretation. This vision encoder is trained to extract high-level visual features from images, al- lowing the model to capture important visual information for accurate interpretation. Veagle distinguishes itself by seam- lessly combining Mistral's exceptional language understanding with the vision abstractor, resulting in a comprehensive model that effectively integrates both textual and visual information. The proficiency of Mistral in language comprehension signifi-cantly enhances Veagle's overall performance. Our methodol-ogy places strong emphasis on the use of a meticulously cu-rated dataset, carefully selected for both pre-training and fine-tuning stages. This dataset serves as a foundation for shap-ing the model's understanding, ensuring robust generalization across different scenarios. Our results show that Veagle has a better grasp of understanding text within images. This is backed up by its impressive performance on standard Visual Question Answering (VQA) tests. Veagle not only outperforms exist- ing models but also establishes a new benchmark for accuracy and efficiency. In conclusion, Veagle represents a cutting-edge model that not only incorporates advanced components but also benefits from the enriching inclusion of curated open sources data, making it a pioneering solution in the evolving landscape of multimodal AI research.

The rest of the paper is organized as follows. Section 2 presents the literature review. 3 highlights the proposed architecture and section 4 includes details of the experiments performed and discusses the results. This is followed by the conclusion in Section 5.

## 2. Literature Survey

In this section, we delve into the related work on large language models and multimodal large language models.

### 1. LLM

Language models (LLMs) have revolutionized the field of natural language processing (NLP), providing capabilities ranging from text prediction to generating coherent and contextually relevant text. In the ever-evolving realm of natural language processing, Large Language Models (LLMs) have undergone a fascinating journey, leaving an indelible mark on the field. The early contributions of trailblazing models like GPT-2[5] and BERT[6] acted as pillars, demonstrating the immense potential that arises from training on vast web-scale text datasets. These models not only laid the groundwork for Natural Language Processing (NLP) but also served as catalysts for subsequent advancements. Among the notable milestones is the monumental GPT-3[7], a model that not only shattered size records but also showcased unparalleled performance in in tackling intricate challenges. With a staggering 175 billion parameters, GPT-3[7] emerged as a powerhouse, excelling in a diverse array of language tasks. Its introduction prompted a re-examination of the limits of model size and sparked renewed interest in the applications and challenges inherent in handling colossal language models. The journey did not conclude with GPT-3[7]; instead, subsequent models like GPT-4[8] and companions like Megatron-turing NLG[9], PaLM[10], Gopher[11], Chinchilla[12], OPT[13], and BLOOM[14] emerged, pushing the boundaries even further. These models, each with unique architectures, training methodologies, and applications, contribute to a dynamic tapestry of research in the expansive domain of large language models. This diversity underscores the ongoing efforts to optimize performance, efficiency, and generalization across an array of linguistic tasks. Recent strides in LLMs have been marked by a nuanced focus on refining models to seamlessly align with human instructions and feedback. Pioneering models such as InstructGPT [15], ChatGPT[16], and the latest iteration, GPT-4[8], stand out as exemplars in this regard. They possess the ability to engage in dynamic, contextually rich conversations, skillfully respond to user prompts, and demonstrate proficiency in intricate tasks such as code generation. These subsequent advancements in LLMs led to the emergence of multimodal large language models, which sought to integrate visual information into the text-based language models This emphasis on harmonizing LLMs with human interaction and instruction signifies a pivotal step toward their practical deployment and integration into real-world applications.

### 2. Multimodal Large Language Models (MLLMs)

In the dynamic landscape of multimodal language models (MLLMs), a paradigm shift is evident as researchers harness the prowess of Large Language Models (LLMs) to transcend traditional linguistic boundaries. Building upon the foundations laid by VisualGPT [17], Frozen [18], Flamingo [19], BLIP2 [20], and other pioneering studies, MLLMs have evolved to proficiently tackle an expanding spectrum of vision-language tasks. These tasks include image captioning, visual question answering (VQA), and bounding box generation, showcasing the robust visual grounding capability inherent in MLLMs. Notably, recent endeavors such as InstructBLIP [3], LLAVA [21, 22], mPlugOwl [2], and BLIVA actively contribute to diversifying the repertoire of tasks that MLLMs adeptly address. Beyond the conventional scope, ongoing research delves into the realm of multimodal instruction tuning, with endeavors like LLaVA[21], InstructBLIP[3], Otter[23], mPLUG-Owl[2] and LLaVA-1.5[22] pioneering advancements in this domain. Despite the ongoing exploration of model architecture and training pipelines, the landscape remains open for innovative solutions. The integration of multimodal information into language models has brought about significant advancements in their performance, efficiency, and generalization across various linguistic tasks.

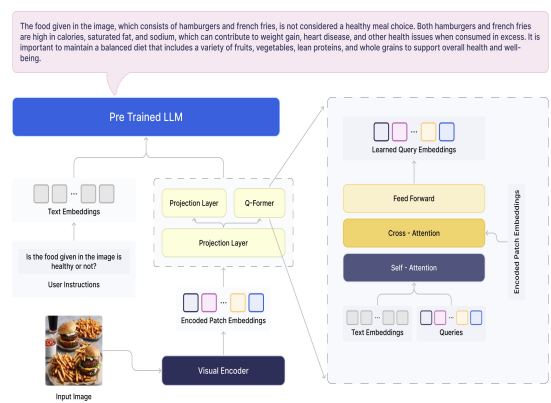

Figure 1: *Veagle Model Architecture: The visual abstractor is responsible for extracting instruction-aware visual features from the output embeddings of the frozen image encoder. Subsequently, these visual features are provided as soft prompts to the frozen Language Model (LLM). The model is then fine-tuned with the language modeling loss to generate the desired response.*

## 3. Proposed Framework

### 1. Architecture Overview

#### 1.1. Image Encoder

A visual encoder is a crucial component of a multimodal models. Visual encoders help the model to extract meaningful representations from visual data. This enables the model to understand the semantics and context of the images, which is important for making accurate predictions or generating relevant outputs. In our experiments, we have adopt a vision encoder(ViT- L/14[24]) from mPlugOwl[2] . This encoder is responsible for extracting meaningful representations from the input images. mPlugOwl[2] has used a novel training paradigm that incorporates a trainable visual encoder, while maintaining the pre-trained language model in a frozen state. This approach enables the model to effectively capture both low-level and higher semantic visual information and align it with the pre-trained language. They have utilize the image-caption pairs from sev-eral datasets, including LAION-400M[25], COYO-700M[26], Conceptual Captions[27] and MSCOCO[28]. model without compromising its performance.

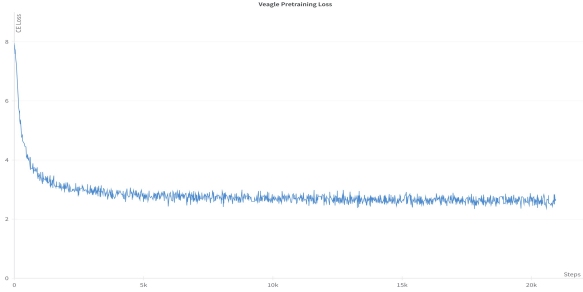

Figure 2: *Pre-training Loss Insights*

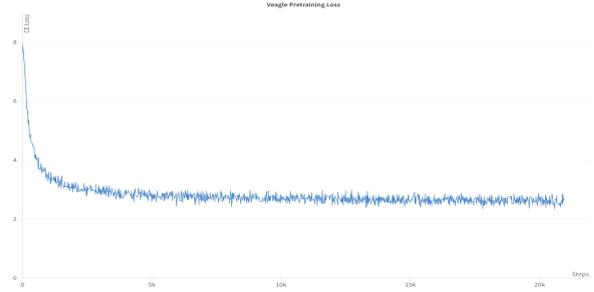

Figure 3: *Fine-tuning Loss Insights*

### 1.2. Visual Abstractor

A visual abstractor serves as a bridge between the visual encoder and the language decoder, enabling the model to effectively process and utilize visual information alongside text, leading to more powerful and versatile multimodal models. It focuses on extracting essential visual features from the en- coded image representations obtained by the image encoder. Large Language Models (LLMs) undergo pretraining primar- ily on textual corpora, presenting a limitation in their innate ability to process image features extracted from Vision En- coders. Addressing this gap, the introduction of the QFormer module in BLIP-2[20] emerged as a critical intermediary, serv- ing to establish a bridge between Vision Encoder and Language Model. Then came BLIVA[1], a groundbreaking combination of BLIP2[20] and LLaVA[22]. However, a linear projection layer have very limited capability in capturing all the informa- tion required for LLM. To overcome the limitations of projec- tion layers in capturing all the necessary information for LLM, we have introduced a multi layer perceptron along with Q- former[20]. In particular, 1 illustrates that our mode generates the embeddings from vision encoder and the output is passed through the projection layer to the Q-former and the second pro- jection layer. The output from the QFormer[20] and Projection layer is concatinated and passed to the LLM which enable better alignment between vision encoders and language models.

### 1.3. LLM

At the heart of multimodal large language models is the Large Language Model (LLM), which serves as the keystone. It takes in instructions and aligned image features, processing this in- formation to generate corresponding answers. In our research, we leverage the capabilities of the many different robust open- source large language models ultimately settling on Mistral[4] due to its superior performance. Mistral 7B surpasses the per- formance of the leading open 13B model (Llama 2[29]) across all benchmarks and outperforms the best released 34B model (Llama 1[29]) specifically in reasoning, mathematics, and code generation tasks. Mistral achieves faster inference through the innovative use of grouped-query attention (GQA) and effec- tively manages sequences of arbitrary length with reduced in- ference cost by incorporating sliding window attention (SWA). This combination of advanced techniques positions Mistral 7B as a leading model in the domain, setting new standards for both accuracy and computational efficiency.

### 2. Training Scheme

The training scheme consists of two stages: Pretraining and Fine-tuning. Figure 4 ilustrate our training paradigm.

#### 2.1. Stage 1: Pre-training

1. In this crucial pre-training stage, the Large Language Model (LLM) is aligned with a visual encoder using image-text pairs from image captioning datasets, facilitating a comprehensive understanding of visual content. The focus is on training the projection layers, refining the mapping of visual and textual information. Throughout this phase, the Vision Encoder, Q- former, and LLM remain frozen, preserving their pre-existing knowledge for subsequent fine-tuning.

#### 2.2. Stage 2: Finetuning

Following pre-training, the Large Language Model (LLM) gains familiarity with the visual embedding space, allowing it to generate image descriptions. However, it lacks the abil- ity to understand finer image details and respond effectively to human queries. In this work, we collect publicly available datasets, COCO, TextCaps, VQAv2, OK-VQA, AOK-VQA, GQA, OCR-VQA, TextVQA, VIzWiz and our in-house curated data. During this phase, the Large Language Model (LLM) and Vision Encoder remain in a frozen state, while the remainder of the model undergoes fine-tuning.

## 4. Experimental Overview

### 1. Datasets

For datasets featuring single-word answers, we adopted an in- novative approach by expanding these responses into detailed and nuanced answers utilizing the advanced capabilities of GPT-4[8] and Mixtral[30]. This strategic enhancement con- tributed to the overall effectiveness of our model, ensuring a more robust and comprehensive understanding of various query types. Addressing the challenge of repeated questions present in certain datasets, we took proactive measures to enhance the diversity and quality of our training dataset. By generating vari- ous different questions that incorporated a wide range of distinct questions, we effectively mitigated redundancy and enriched the training dataset, thereby fostering improved generalization and performance in handling diverse queries. This meticulous pro- cess of dataset augmentation and refinement played a pivotal role in optimizing the overall performance and reliability of our model. The careful compilation, filtering, and augmentation of diverse datasets played a crucial role in maximizing the perfor- mance and reliability of our model.

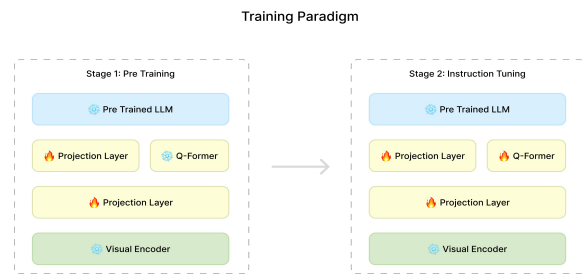

Figure 4: *Overview of Veagle training paradigm*

Table 1: *Performance of the proposed model for different opensourced datasets.*

|              | Veagle | BLIVA | InstructBLIP | mPlugOwl | LLAVA |
|--------------|--------|-------|--------------|----------|-------|
| ok_vqa       | **49.3** | 43.4 | 30.8 | 34.1 | 46.2 |
| ocr_vqa      | 48.3 | 38.5 | 32.1 | 61.4 | **67.2** |
| scienceQA    | **58.1** | 16.1 | 40.2 | 51.8 | 56.5 |
| coco_caption | 57.9 | 56.4 | 51.2 | 55.6 | **62.7** |
| ai2diagram   | **56.3** | 50.8 | 31.9 | 48.5 | 50.9 |
| chart_qa     | **13.4** | 13.2 | 3.4 | 10.2 | 3.1 |
| gqa          | **44.2** | 28.6 | 40.8 | 33.9 | 43.9 |
| text_vqa     | 22.5 | 23.1 | 20.5 | 32.6 | **37.2** |

## 2. Results

Our experimental results demonstrate the effectiveness of our approach, with significantly improved performance across vari-ous datasets.

### 2.1. Baseline vs Proposed Protocol

We used four advanced baseline models BLIVA[1], instructBLIP[3], mPlugOwl[2], and LLAVA[22] for our analysis. For each of these models, we took an image and a question, input them into the model, and noted down the response it gave. To evaluate the precision of the provided responses, we employed GPT-4[8] as our assessment model. This model categorized the answers into two distinct classi- fications: either correct or incorrect. The accuracy outcomes corresponding to each dataset for various different models, obtained through the utilization of this evaluation method, are comprehensively presented in Table 1. Our proposed model achieved an impressive level of accuracy when compared to other open sourced baseline models.

### 2.2. In-House Test Dataset

To assess how well our model performs in different scenarios and if it generalizes effectively, we created an in-house test dataset. This dataset comprises various types of tasks, includ- ing captioning, optical character recognition (OCR), general vi- sual question-answering (VQA), technical VQA, and reasoningVQA. Importantly, our model has never encountered this spe- cific dataset during its training process. Subsequently, we con- ducted thorough evaluations of all the models using this test dataset, and the outcomes are promising. Detailed results are presented in Table 2.

Table 2: *Performance of our proposed model Veagle for our in-house test dataset.*

|           | Veagle | BLIVA | InstructBLIP | mPlugOwl | LLAVA |
|-----------|--------|-------|--------------|----------|-------|
| Test Data | **76.4** | 63.1 | 59.3 | 68.6 | 66.5 |

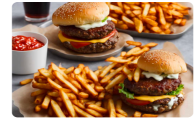

Figure 5: *Qualitative examples produced by our Veagle model showcase a spectrum of its diverse capabilities. These demon- strations include intricate visual scene understanding and rea-soning, multi-turn visual conversation, and more.*

### 2.3. Qualitative Ananlysis

In this section, we present the qualitative outcomes derived from our assessment set. This set of evaluations was carefully curated to analyze the model's performance on intricate and challeng- ing tasks. The tasks were selected and collected by our team for the purpose of understanding the model's effectiveness beyond numerical measures, delving into the nuanced aspects of its per- formance. Figure 5 is showing the effectiveness of our model. More examples are given in 7

## 5. Conclusion

In conclusion, the Veagle multi-modal model stands out as a formidable contender, consistently outperforming established benchmarks in diverse domains. Through the strategic fusion of various modules curated from extensive research, Veagle show- cases remarkable performance, not only meeting but exceedingthe expectations set by existing models. However, our work also reveals areas that still require refinement, emphasizing the on-going nature of the pursuit for perfection. This acknowledgment underscores the need for further exploration and optimization, recognizing that the path to excellence in multi-modal models like Veagle continues to unfold. As we navigate this landscape, Veagle remains a promising catalyst for future advancements in Vision-Language Models, beckoning further investigation and innovation in this dynamic field.

## 7. Appendix

### 1. GitHub

The code repository, along with detailed documentation, can be found at `https://github.com/superagi/Veagle`.

### 2. Huggingface

The Veagle model, along with detailed documentation is available at `https://huggingface.co/SuperAGI/Veagle`

### 3. Training Parameters

Table 3: *Training parameters in both the stages*

|  | Epochs | Optimizer | l-rate | Batch size | Weight decay |
| --- | --- | --- | --- | --- | --- |
| Pre-training | 3 | AdamW | 1e-5 | 8 | 0.05 |
| Fine-tuning | 2 | AdamW | 1e-5 | 10 | 0.05 |

### 4. Compute complexity

We have used 8 NVIDIA A100 with a batch size of 10 for both pre-training and fine-tuning. For inference 1 NVIDIA A6000 is used.

### 5. Qualitative Examples

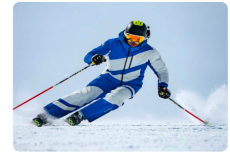

Describe the image.

A man in a blue and white ski suit is skiing down a snowy slope.

What is the use of ski suit a man is wearing?

The ski suit is designed to provide warmth, protection, and comfort while skiing down the hill. It is made of materials that are breathable, waterproof, and insulated to keep the skier warm and dry during their time on the slopes. Additionally, ski suits often have features such as zippers, vents, and adjustable cuffs to help the skier regulate their body temperature and adapt to changing weather conditions.

What are the risk involve while skiing?

Skiing involves the risk of falling, which can result in injuries such as sprains, strains, fractures, or head injuries. Skiers may collide with other skiers, trees, rocks, or other obstacles, leading to injuries or accidents. Adverse weather conditions, such as heavy snowfall, fog, or strong winds, can make skiing more challenging and increase the risk of accidents. Skiing at high altitudes can lead to altitude sickness, which can cause symptoms such as headache, dizziness, nausea, and shortness of breath. Prolonged exposure to cold temperatures can lead to frostbite and hypothermia, which can be life-threatening if not treated promptly.

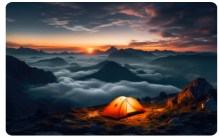

What sort of activites can one enjoy at such place?

At a place like this, one can enjoy various outdoor activities such as hiking, camping, fishing, and birdwatching. The presence of a tent suggests that camping is an option, allowing visitors to spend time in the wilderness and connect with nature. Hiking can be a great way to explore the surrounding terrain and take in the breathtaking views of the mountains and valleys. Fishing enthusiasts can try their luck at catching fish in nearby bodies of water, while birdwatchers can observe various bird species that inhabit the area. Additionally, photography enthusiasts can capture stunning images of the scenery, especially during sunrise or sunset when the colors are more vibrant. Overall, such a place offers a wide range of activities suitable for different interests and preferences, making it an ideal destination for nature lovers and outdoor adventurers.

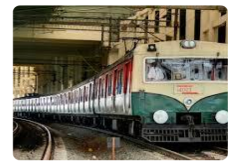

What kind of transportation system is displayed in the image?

The image displays a train transportation system.

Where is the train currently?

The image displays that a train is at train station, which is a type of transportation system where trains stop to pick up and drop off passengers.

Figure 6: *Examples generated by our Veagle model exemplify a broad spectrum of its diverse capabilities. These showcases encompass intricate visual scene comprehension and reasoning, multi-turn visual conversation, and various other impressive functionalities.*

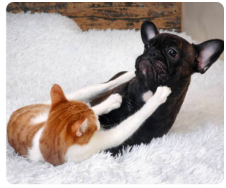

Write a poem that rhymes very well based on the above image

A dog and a cat, side by side,In a cozy blanket, they rest and hide.Their soft snores echo in the room,A peaceful slumber, a tranquil bloom.

Their paws intertwined, a gentle sign,Of friendship and love, a bond divine.They dream of adventures, far and wide,In their dreams, they soar, they glide and hide.

In the morning, they'll wake up and play,Exploring the world, in their own unique way.But for now, they sleep, content and at ease,In the warmth of each other's company and peace.

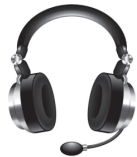

Help me write an advertisement for an ad campaign based on above image

The ad campaign for a headset would emphasize the high-quality sound and noise cancellation features, as well as the sleek design and comfortable fit. The tagline could be something like "Experience crystal-clear sound with our noise-cancelling headset" or "Immerse yourself in sound with our advanced headset technology." The ad could also showcase the headset in use, such as a person wearing it while working on a computer or listening to music, to demonstrate its versatility and practicality. Additionally, the ad could include testimonials from satisfied customers, highlighting their positive experiences with the headset. Overall, the ad campaign would aim to position the headset as a must-have accessory for anyone who values high-quality sound and wants to minimize distractions in their environment.

Figure 7: *From pixels to pitches, our model's ability to generate poems, ads, and songs from images is a testament to its multi-dimensional creative prowess.*

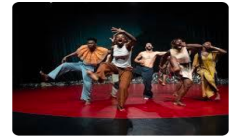

what are the people doing in this image?

The people in the image are performing a dance routine on a stage. They are dressed in various costumes, including white dresses, and they are dancing together in a coordinated manner. The stage is decorated with red carpets, which adds to the overall aesthetic of the scene. The dancers appear to be engaged in a lively and energetic performance, showcasing their skills and creativity through their choreographed movements. Overall, the image captures a moment of artistic expression and collaboration between the dancers as they entertain the audience with their dance routine.

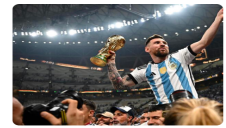

What might be the reason of celebrations in the image?

The reason for the celebrations in the image could be that the soccer team has just scored a goal or won a match. The players are seen raising their arms in the air and cheering, indicating their excitement and satisfaction with the outcome of the game. It is also possible that they are celebrating a personal achievement or milestone within the team, such as a player's birthday or reaching a significant career milestone. Regardless of the specific reason, the players are expressing their joy and camaraderie through their celebratory actions.

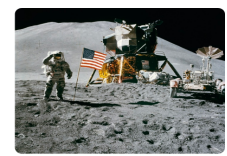

What is the significance of this moment in history?

The image captures a significant moment in history when astronauts Neil Armstrong and Edwin "Buzz" Aldrin became the first humans to set foot on the moon during the Apollo 11 mission on July 20, 1969. This event marked a major milestone in space exploration and human achievement, as it demonstrated humanity's ability to travel beyond Earth's atmosphere and land on another celestial body

Figure 8: *Examples generated by our Veagle model showing its innovative capabilities.*